\pdfoutput=1

\documentclass[11pt]{article}

\usepackage[final]{acl}

\usepackage{times}
\usepackage{latexsym}

\usepackage[T1]{fontenc}

\usepackage[utf8]{inputenc}

\usepackage{microtype}

\usepackage{inconsolata}

\usepackage{graphicx}

\usepackage{amsmath, amsfonts, amssymb}
\usepackage{algorithmic}
\usepackage{algorithm}
\usepackage{multirow}
\usepackage{color,soul}
\usepackage{csquotes}

\title{Media Framing through the Lens of Event-Centric Narratives}

\author{
   Rohan Das$^{1}$ \,\,
   Aditya Chandra$^{1}$ \,\, 
   I-Ta Lee$^{2}$ \,\,
   Maria Leonor Pacheco$^{1}$ \\
   $^1$University of Colorado Boulder \,\, $^2$Independent Researcher\\
   $^1$\texttt{\{rohan.das, aditya.chandra, maria.pacheco\}@colorado.edu} \\
}

\begin{document}
\maketitle


\begin{abstract}
From a communications perspective, a frame defines the packaging of the language used in such a way as to encourage certain interpretations and to discourage others. For example, a news article can frame immigration as either a boost or a drain on the economy, and thus communicate very different interpretations of the same phenomenon. In this work, we argue that to explain framing devices we have to look at the way narratives are constructed. As a first step in this direction, we propose a framework that extracts events and their relations to other events, and groups them into high-level narratives that help explain frames in news articles. We show that our framework can be used to analyze framing in U.S. news for two different domains: immigration and gun control.

\end{abstract}

\section{Introduction}

Framing involves curating certain aspects of issues or events and coherently organizing them in a way to make arguments, with the goal of promoting a particular interpretation, evaluation or solution~\citep{entman-2003}. For example, a news story about immigration could be framed as a crisis of illegal border crossings, or it could be framed as a search for better opportunities by people fleeing violence and poverty. Similarly, debates about gun control often involve projecting guns as either instruments of violence or tools of self-defense.

\begin{figure}
    \centering
    \includegraphics[width=1\columnwidth]{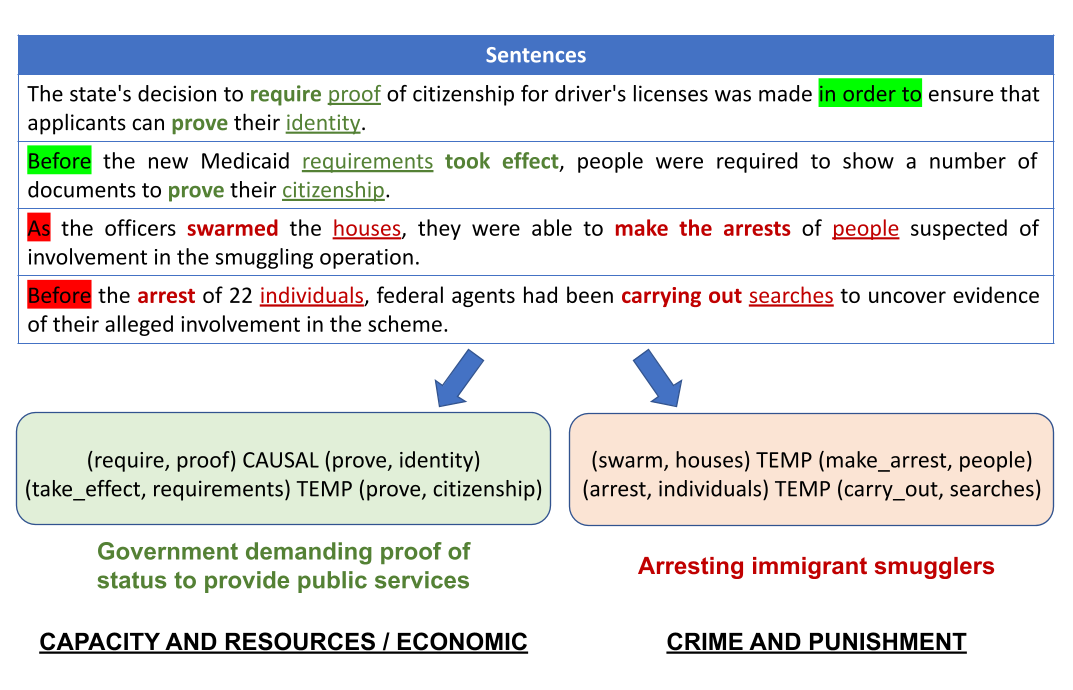}
    \caption{Motivating example for grouping narratives. \textbf{Verbs} are in bold. \underline{Objects} are underlined. \hl{Relations} are highlighted. Colors indicate narrative clusters. Capitalizations indicate \citet{Boydstun2014} policy frames.}
    \label{fig:motivating_example}
\end{figure}

Media framing analysis is essential for understanding how public opinion is formed and how social movements gain momentum. By examining the ways in which different actors frame issues, we can gain insights into the underlying power dynamics at play and the strategies used to persuade and mobilize people. Moreover, framing analysis can help us to identify and challenge harmful stereotypes and biases that perpetuate inequality and injustice.

Dominant computational approaches to media framing rely on high-level topic markers to conceptualize frames~\citep{ali-hassan-2022-survey}, either by manually constructing topical taxonomies~\cite{Boydstun2014,card-etal-2015-media,liu-etal-2019-detecting} or by extracting latent semantic structures using topic models~\citep{DiMaggio2013ExploitingAB, Gilardi2020PolicyDT}. The main drawback of these approaches is that the resulting categories can be too broad to understand a frame's nuances. By reducing frames to a few co-occurring keywords (e.g., city, building, park, downtown) or to broad topics (e.g., economic, politics), we can fail to capture \textit{how} different aspects are chosen and organized to make an argument~\cite{entman-2003, doi:10.1177/1742715005051857}. 

As an example, consider the policy frame taxonomy proposed by \citet{Boydstun2014}, where framing dimensions correspond to broad themes like ``economic'', ``crime'' and ``capacity and resources''. Under the same ``economic'' marker, a news article can frame immigration as either a boost or a drain on the economy. The author can either argue that immigrants contribute to economic growth by filling labor shortages and starting businesses, or contend that immigrants compete with citizens for jobs and drive down wages.



In this paper, we propose a new approach to media framing analysis that centers the role of narratives. \citet{castricato-etal-2021-towards} define narratives as stories that convey information, shape perceptions, and influence attitudes and behaviors. Our main goal with this approach is to find repeating story-telling patterns that can help disambiguate and explain high-level framing dimensions. For example, when framing immigration as a crime issue, journalist may resort to telling stories about illegal smuggling and how it results in raids and arrests.

Computational approaches to narrative analysis largely follow a model where narratives are considered to be sequences of events that unfold over time, involving characters, settings, plots, and are often characterized by their temporal structure and causal relationships \citep{piper-etal-2021-narrative}. We build on this body of work, and propose a framework that extracts event-centric narrative representations and groups them into higher-level themes that help explain broad frames. To do this, we first extract $(verb, object)$ events from open text. Then, for every pair of events we predict whether they are \textit{temporally} related (i.e., do they occur in chronological order?) or \textit{causally} related (i.e., are they involved in a cause-and-effect relationship?). Finally, we cluster $(event, relation, event)$ chains into higher-level narratives that are informative for predicting the policy frame taxonomy proposed by \citet{Boydstun2014}.




To illustrate this, consider the example outlined in Fig.~\ref{fig:motivating_example}. Here, we observe that triplets extracted from news articles about immigration such as \textit{((require, proof), CAUSAL, (prove, identity))} and \textit{((take\_effect, requirements), TEMP, (prove, citizenship))} can be grouped into the broader theme of ``the government requiring proof of status to provide public services'', which in turn can be tied to \citet{Boydstun2014} policy frames like ``capacity and resources'' or ``economic''. 


We make the following contributions: (1) We propose a computational framework to study media framing through the lens of event-centric narratives. (2) We demonstrate the generalizability of our framework by applying it to two different news domains: immigration and gun control. (3) We perform a comprehensive evaluation and show that we can produce high-quality narrative clusters for the immigration domain, and that the induced clusters provide significant signal for predicting and explaining the \citet{Boydstun2014} policy frame taxonomy for both domains.

\section{Related Work}

The related work can be organized in two main streams: \textit{Computational Framing Analysis} and \textit{Narrative Representations}.

\paragraph{Computational Framing Analysis}
A popular family of framing analysis methods adopts unsupervised techniques such as topic modeling to identify latent themes \citep{DiMaggio2013ExploitingAB,nguyen2015,Gilardi2020PolicyDT}. However, these methods are limited in their ability to capture the nuances of framing. The results of topic models are usually a list of keywords and their interpretation is usually unaligned with the detailed aspects of framing \citep{ali-hassan-2022-survey}. 

Supervised learning \citep{johnson-etal-2017-leveraging, khanehzar-etal-2019-modeling, 10.1145/3394231.3397921, huguet-cabot-etal-2020-pragmatics, mendelsohn-etal-2021-modeling} and lexicon expansion \citep{field-etal-2018-framing, roy-goldwasser-2020-weakly} techniques have also been applied to analyze framing. For these methods to work, a concrete taxonomoy of relevant frames and their representations are required. However, the manual construction of such taxonomies is time-consuming and is not generalizable across different domains. Moreover, they suffer from the same limitation as topic modeling in terms of capturing the nuances of framing. 

\citealp{khanehzar-etal-2021-framing} proposed a semi-supervised interpretable multi-view model for identifying media frames. The model jointly learns dense representations for events and actors, which are then integrated with a latent semantic role representation to predict media frames of documents. However, this method heavily relies on local information, which is a significant limitation. The model fails to incorporate global context, often mislabeling the primary frame of related articles. For example, the \textit{Political} frame is often misclassified as \textit{Legality} due to the significant overlap in keywords.


\paragraph{Narrative Representations}

\citealp{chambers-jurafsky-2008-unsupervised} introduced an unsupervised method for learning narrative event chains from raw newswire text. Narrative chains, as defined in their work, are sequences of events that share a protagonist as the event actor and contribute to a coherent narrative. Their method involved identifying events within text using syntactic analysis, determining their temporal order based on co-occurrence patterns and grammatical relationships, and clustering related events into coherent chains. To evaluate the quality of the learned event chains, the authors introduced two evaluation tasks: narrative cloze and order coherence. Their work laid the foundation for subsequent research on event sequence modeling and story generation, demonstrating the feasibility of unsupervised learning for complex narrative structures.

\citealp{lee-etal-2020-weakly} presented a weakly supervised method for learning contextualized event representations from narrative graphs. By representing events as nodes and typed relationships as edges in these graphs, they were able to capture the global context of the narrative. These representations can then be used to effectively identify discourse relations in extrinsic evaluations.
\citealp{Zhang2021SalienceAwareEC} combined salience identification \citep{liu-etal-2018-automatic, jindal-etal-2020-killed} and discourse profiling techniques \citep{choubey-etal-2020-discourse} to isolate the main event chains from less relevant events. They constructed temporal relation graphs from documents and applied various filtering levels to the extracted events. By traversing the directed edges in the filtered graph, they extracted linear event chains. The resulting event chains were used to build event language models, which were then evaluated on story cloze and temporal question answering tasks. \citealp{hatzel-biemann-2023-narrative} proposed a novel approach to narrative modeling using narrative chain embeddings and explored applications to a down-stream task in the form of replicating human narrative similarity judgments. 

Recent work adopts pre-trained language models to further advance narrative representations. \citealp{10.1145/3397271.3401173} modeled event elements by fine-tuning a masked language model on event chain representations. \citealp{li-etal-2020-connecting} used an autoregressive language model to learn event schemas from salient paths in an event-event relation graph.


\section{Extracting Narratives}\label{sec:framework}

This section describes our framework to extract event mentions, their relations to obtain narrative chains, as well as our approach to cluster narrative chains into high-level themes. 

\subsection{Extracting Events}

In this work, we take a \emph{verb-centric} view of events. Particularly, we follow the widely adopted event representation consisting of a pair of a dependency type (e.g., subject or object) and predicate tokens (e.g., verb) \citep{GranrothWilding2016WhatHN}. 

To extract event mentions from documents, we adopt the ETypeClus framework \citep{shen-etal-2021-corpus}. In this framework, an event mention consists of a verb and its corresponding object in a given sentence. To extract verb and object heads in sentences, we use a dependency parser\footnote{We use the Spacy \texttt{en\_core\_web\_lg} model.} to obtain the dependency parse tree of each sentence and select all non-auxiliary verb tokens\footnote{A token with part-of-speech tag \texttt{VERB} and dependency label not equal to \texttt{aux} and \texttt{auxpass}.} as our candidate verbs. We then identify the corresponding object head for each candidate verb depending on whether the sentence is in active or passive voice. We then process the entire corpus of documents to extract a list of all the $(verb, object)$ mentions in each document.

\subsection{Extracting Relations}

To extract relations, we build a classifier to predict relations between each pair of extracted events in a given document. We focus on two types of relations: \textit{temporal relations} -- the chronological relationship between events, and \textit{causal relations} -- the cause-and-effect relationships between events.

To do this, we create a comprehensive training dataset from ASER (Activities, States, Events and their Relations)~\cite{10.1145/3366423.3380107}, a large-scale eventuality knowledge graph that contains 14 relation types taken from the Penn Discourse TreeBank~\cite{prasad-etal-2008-penn}, as well as a \emph{co-occurrence} relation. In total, ASER contains 194-million unique eventualities and 64-million unique edges among them. Relations are defined as triplets ($e_h, r, e_t$), where $e_h$ and $e_t$ are head and tail events and $r$ is the relation type. The head and tail events are sentences that follow a syntactic pattern (e.g., subject-verb-object). For our dataset, we retain only verbs and objects. For example, if $e_h= (am, hungry)$ and $e_t= (eat, pizza)$, then relation $r= Result$.

We consider only a subset of relations in ASER for building our training dataset. To choose this subset, we use two threshold criteria: (1) the relation must appear in at least five different unique event pairs, and (2) if more than one relation exists between two events $E_h$ and $E_t$, we take the one with the maximum strength. To calculate the strength, we use the one-hop relation retrieval inference score, shown in equation \eqref{1}:

\begin{equation}
\label{1}
    P(r|E_h,E_t) = \cfrac{f(E_h,r,E_t)}{\Sigma_{r'\in R} \;f(E_h,r',E_t)}
\end{equation}

\noindent where $R$ is the relation set, and $f(E_h,r,E_t)$ is the number of times the triplet appears in the knowledge base. A higher score indicates a stronger belief that $r$ is the correct relation for the given entity pair, making it a probabilistic measure for selecting the most likely relation type. Additional pre-processing details are included in App. ~\ref{sec:aser_preprocess}. 



We retain the five most common PDTB relation types: Precedence, Succession, Synchronous, Reason, and Result. We group these into two categories, \textit{temporal} (Precedence, Succession, Synchronous) and \textit{causal} (Reason, Result). To handle the absence of a relation between events, we create negative examples using all discarded relation types. Tab. \ref{tab:rel_data} summarizes the resulting dataset. 

\begin{table}
\centering
\begin{tabular}{ccc}
\hline
 \multicolumn{1}{c}{\textbf{Temporal}} & \multicolumn{1}{c}{\textbf{Causal}} & \multicolumn{1}{c}{\textbf{None}} \\ \hline
 52,556                                 & 35,827                               & 212,555                            \\   \hline                
\end{tabular}
\caption{Class counts in the training data for the relation classification module.}
\label{tab:rel_data}
\end{table}

\begin{figure}[t]
    \centering
    \includegraphics[width=\columnwidth]{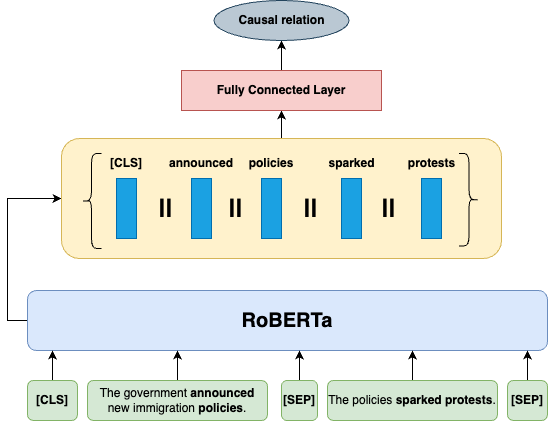}
    \caption{Relation Classifier Architecture}
    \label{fig:arch}
\end{figure}

We outline the architecture of our relation classifier in Fig. \ref{fig:arch}. We use pre-trained RoBERTa ~\cite{liu2019robertarobustlyoptimizedbert} to represent the event verb-object pairs, as well as the sentences that they appear in, and add a one-layer classifier on top. All parameters are fine-tuned during training. For further implementation details refer to App. ~\ref{sec:rel_exp_details}.  


Finally, we use the relation classifier to predict a relation (or the absence of one) between each pair of extracted events in a given document. Only those event pairs with a temporal or causal relation are retained and used to construct a set of narrative chains for each document in the corpus. We only consider single-hop chains, and represent them as $(event_{1}, relation, event_{2})$, where each event is a $(verb, object)$ pair.

\begin{table*}[ht!]
    \centering
    \resizebox{\textwidth}{!}{%
    \begin{tabular}{ccccccc}
    \hline
    \textbf{Issue}       & \textbf{Training Set} & \textbf{Test Set} & \textbf{Unique Events} & \textbf{Narrative Chains} & \textbf{Avg. chains per article} & \textbf{Frame Labels} \\ \hline
    \textbf{Immigration} & 1,772                 & 197               & 14,965                 & 108,348                   & 54                               & 15                    \\ 
    \textbf{Gun Control} & 1,773                 & 198               & 14,134                 & 113,602                   & 57                               & 14                    \\ \hline
    \end{tabular}%
    }
    \caption{Summary of the dataset subsets used from the Media Frames Corpus, along with the number of extracted events, narrative chains, the average number of narrative chains per article, and the total number of frame labels in each dataset. The 'Other' frame label does not appear in our subset of the gun control dataset.}
    \label{tab:dataset-summary}
\end{table*}

\subsection{Clustering Narrative Chains}

We cluster narrative chains to identify distinct narratives themes and constructs in the documents. This enables us to capture the nuanced aspects that were chosen and organized to make certain arguments, as opposed to relying on broad topics derived from clustering just words or standalone events. This approach also allows us to cut through the noise and focus on the most salient narratives in a document, without sacrificing global context, thus resulting in a richer and concise representation of the document. 

We adopt an LLM guided clustering method that allows us to abstract away from the \emph{(event, relation, event)} chains to a high-level, in-context textual representation. We prompt an instruction fine-tuned Llama 3.1 8B model \citep{Dubey2024TheL3} in a zero-shot setting (see Sec.~\ref{sec:prompts} for the prompt template). We provide the full document and a corresponding narrative chain, and prompt the model to expand the narrative chain into a short sentence that describes the causal or temporal sequence of events. Examples of narrative chain expansions are shown in Tab. \ref{tab:llm-expansion}.

Once narrative chains have been expanded, an SBERT model \citep{Reimers2019SentenceBERTSE} that was trained on semantic search tasks, is used to compute sentence embeddings for each of the generated sentences. These sentence embeddings are then used to cluster similar narrative chains together. We use the K-means clustering algorithm to cluster the chains into a fixed number of clusters. We experiment with different numbers of clusters ranging from 25 to 200, in increments of 25. 

\begin{table*}[ht!]
\centering
\resizebox{\textwidth}{!}{%
\begin{tabular}{ll}
\hline
\textbf{Narrative Chain}                         & \textbf{LLM Expansion}                                                                                                                                                                                                          \\ \hline
$((pay, fine), TEMPORAL, (become, resident))$ & \begin{tabular}[l]{@{}l@{}}After paying a fine, illegal immigrants would be able to become \\ permanent residents under the proposed U.S. Senate immigration bill.\end{tabular}   \\

\hline
$((require, check), CAUSAL, (close, loophole))$ & \begin{tabular}[l]{@{}l@{}}The decision to require background checks at gun shows \\ 
was a key factor in closing the so-called "gun show loophole".\end{tabular}    
\\ \hline
\end{tabular}%
}
\caption{Examples of narrative chain expansions generated by prompting a Llama 3.1 8B model.}
\label{tab:llm-expansion}
\end{table*}
\section{Analysis}
This section describes the experiments and quantitative analysis that we perform to evaluate the quality of the narratives, as well as our approach towards explaining framing with the help of these narratives. We also include our findings from a qualitative evaluation of the different narrative themes observed in the narrative clusters across different framing issues.

\subsection{Datasets} 

We performed our experiments and analysis on news articles covering two different domains: immigration and gun control. We take documents from the Media Frames Corpus \citep{card-etal-2015-media}, which consists of annotated news articles across 15 different framing dimensions at both the article level and the text spans that cued them. In this work, we investigate the role of narrative structure in framing analysis by evaluating how narrative chains can be used to predict and explain article level framing labels. We use a subset of the dataset from both domains, and the splits are shown in Tab. \ref{tab:dataset-summary}. We apply our narrative chain framework to the datasets to extract the events, relations and narrative clusters for each news article. 

\subsection{Quality of Narratives} 

In this section, we perform an intrinsic evaluation of the narratives extracted using the framework described in Sec.~\ref{sec:framework}. To do this, we look at the performance of our relation classifier, as well as the quality of the resulting narrative clusters. 

\paragraph{Relation Prediction} 
We evaluate the performance of the relation classifier on our subset of the filtered relation prediction dataset derived from ASER using 5-fold cross-validation. 
We use the AdamW optimizer and a weighted cross-entropy loss function to train our models. All hyper-parameters, experimental setup, and cross-validation results are reported in App.~\ref{sec:rel_exp_details}. To measure performance, we compute accuracy, precision, recall, and F1 for each class, and report the macro averages to account for class imbalance. We compare our model with three baselines. (1) \textbf{Majority Class} - always predicts the majority class, in this case the \emph{None} label. (2) \textbf{Random} - randomly assigns a relation label to each event pair. (3) \textbf{Logistic Regression} - trained using 300-dimensional GloVe embeddings \citep{pennington-etal-2014-glove}. For each event pair, we first compute sentence embeddings for the phrases containing the corresponding event by averaging the GloVe word vectors. We then concatenate these two embeddings into a single feature vector, which serves as the input for the classifier. 
Results are reported in Tab. ~\ref{tab:mclass-f1}.

\begin{table}[ht!]
    \centering
    \resizebox{\columnwidth}{!}{%
    \begin{tabular}{lccccc}
    \hline
            \textbf{Models}   & \multicolumn{1}{c}{\textbf{Temporal}} & \multicolumn{1}{c}{\textbf{Causal}} & \multicolumn{1}{c}{\textbf{None}} & \multicolumn{1}{c}{\textbf{All}} \\ \hline
\textbf{Majority Class}    & $0.00_{0.00}$        & $0.00_{0.00}$     & $0.83_{0.00}$ & $0.27_{0.00}$      \\
\textbf{Random}  & $0.23_{0.01}$       & $0.18_{0.01}$    & $0.45_{0.01}$ & $0.28_{0.01}$      \\
\textbf{Logistic Regression}      & $0.32_{0.01}$       & $0.22_{0.00}$    & $0.51_{0.00}$ & $0.35_{0.00}$      \\
\textbf{Our Model}      & $0.59_{0.01}$       & $0.42_{0.01}$    & $0.78_{0.00}$ & $0.60_{0.01}$      \\ \hline
    \end{tabular}}
    \caption{F1 scores for the multi-class relation prediction model (average and standard deviation over all five folds).}
    \label{tab:mclass-f1}
\end{table} 

We find that our model was able to achieve an average macro-F1 score of 0.6 which is in line with recent work on implicit discourse relation prediction~\cite{yung-etal-2024-prompting}. Unsurprisingly, predicting causal relations is significantly more difficult than predicting temporal relations. We also find that for causal relations, recall is significantly better than precision, which is appropriate for our use case given that we care about achieving high coverage, but we can make up for some degree of noise by aggregating narrative chains in our clustering step. On the other hand, our model is reasonably good at discarding event pairs where no temporal or causal relation occurs. 

\paragraph{Narrative Clustering} We evaluate the quality of the resulting narrative clusters by performing an intrusion test. Given two random samples from the top 25\% of narrative chains from a cluster, we inject a randomly sampled chain from another cluster as a negative example. The narrative chains are ranked based on their distance to the cluster centroid. Two annotators are asked to independently identify the intruder, and a third annotator attempts to resolve conflicts without looking at previous annotations. Intuitively, if the clustering results are clean and capture similar high level narrative patterns, then the annotators will find it easier to identify the intruder. We report the inter-annotator agreement (Krippendorff's alpha) and the intruder labeling accuracy to measure the quality of the generated clusters in Tab.~\ref{tab:intrusion}. 

\begin{table}[h!]
\centering
\resizebox{\columnwidth}{!}{%
\begin{tabular}{cccc}
\hline
\multicolumn{2}{c}{\textbf{Immigration}} &
  \multicolumn{2}{c}{\textbf{Gun Control}} \\ \hline
\textbf{\begin{tabular}[c]{@{}c@{}}Inter-Annotator \\ Agreement\end{tabular}} &
  \textbf{Accuracy} &
  \textbf{\begin{tabular}[c]{@{}c@{}}Inter-Annotator \\ Agreement\end{tabular}} &
  \textbf{Accuracy} \\ \hline
82.61 &
  67.5 &
  65.89 &
  37.5 \\ \hline
\end{tabular}%
}
\caption{Intrusion Test Results: Krippendorff's alpha is used to compute inter-annotator agreement ($\alpha$ = 0 represents random agreement, $\alpha$ = 100 represents perfect agreement). Intrusion labeling accuracy is reported in percentage.}
\label{tab:intrusion}
\end{table}

We observed high inter-annotator agreement for the immigration dataset, as well as good labeling accuracy, indicating high quality clusters, each representing well-defined, semantically coherent themes. However, we observed low labeling accuracy for the gun control dataset (random baseline score of 33\% for 3 intruder candidates). Our annotators noted that the gun control dataset lacks variation in narrative themes which can (1) make our framework more susceptible to noise in the relation extraction step, and (2) result in overlapping clusters, thus making this a comparably harder annotation task.

\begin{figure*}[ht!]
    \centering
    \includegraphics[width=1\textwidth]{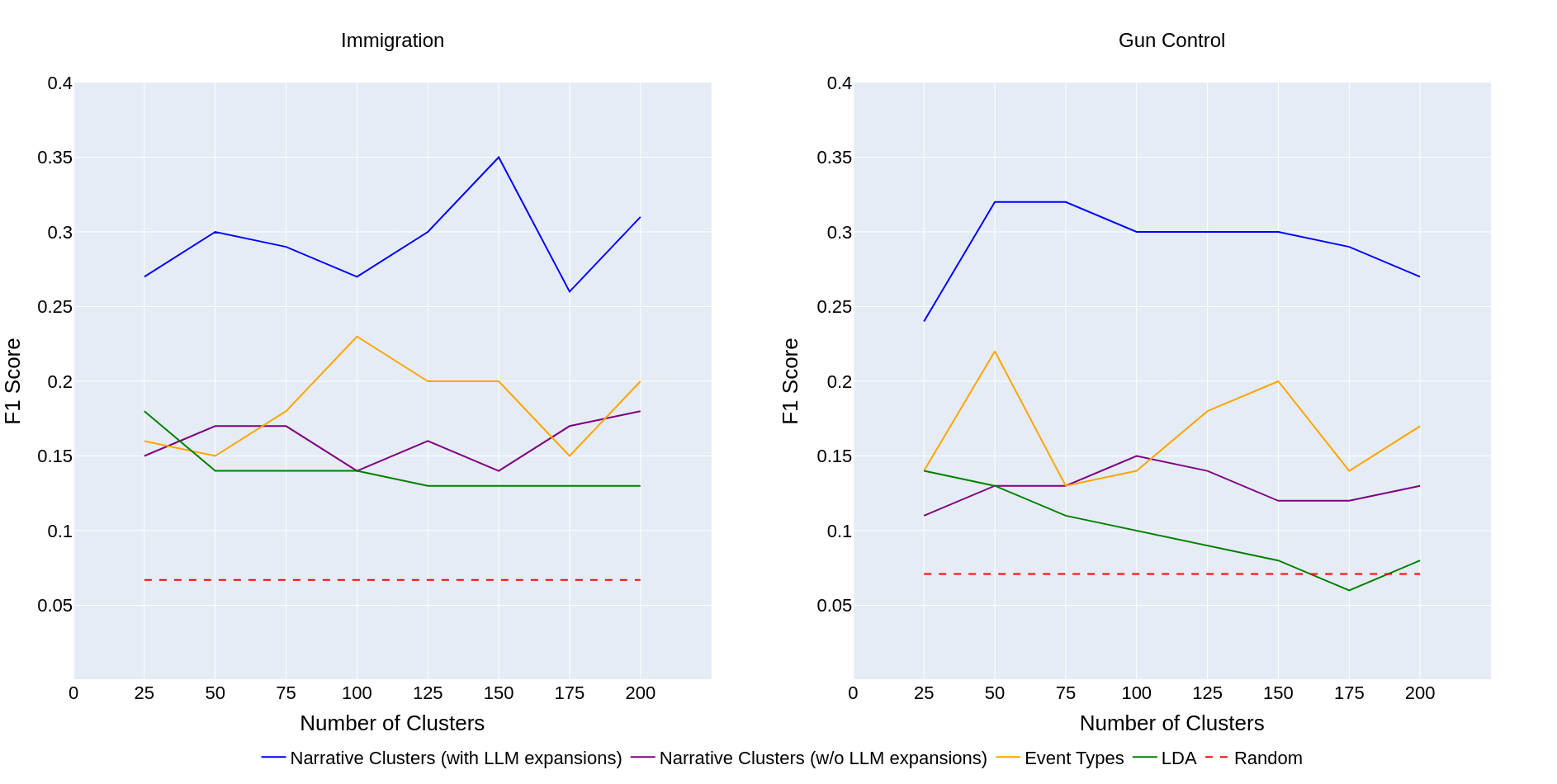}
    \caption{Frame prediction results on the immigration and gun control datasets using \textit{only} cluster features. The model powered by Narrative clusters (with LLM expansions) outperforms four baselines: (1) Random, (2) LDA topics, (3) Event Types, and (4) Narrative Clusters (w/o LLM expansions).}
    \label{fig:frame-prediction-results}
\end{figure*}

\subsection{Explaining Frames with Narratives}

In this section, we explore the potential of our narrative clusters to predict and explain framing dimensions in the \cite{Boydstun2014} policy frame taxonomy. To do this, we first evaluate the predictive signal of the resulting narrative clusters, both in isolation and in addition to textual information. Then, we perform a comprehensive qualitative analysis of the resulting clusters and their relation to the high-level framing dimensions. 

\subsubsection{Frame Prediction} To evaluate whether narrative clusters have any predictive signal for the \citet{Boydstun2014} high-level framing dimensions, we perform the following two experiments.

\paragraph{Narrative Cluster Features} The first experiment attempts to predict article level frames by looking \textit{only} at the latent narrative themes (i.e. clusters) that were identified for a given document. The intuition behind this experiment is not to achieve good prediction performance, as no direct language information is used, but to gauge how much signal is implicitly encoded in the association of a document to the high-level narrative patterns identified. 

To do this, we map each narrative chain in a document $d$ to the cluster it was assigned to. Let $f_k$ represent the frequency of the $k$-th narrative cluster in $d$, defined as the number of narrative chains within that cluster in $d$:
\begin{equation}
\label{2}
f_k = n_k
\end{equation}
where $n_k$ is the count of narrative chains in the $k$-th cluster. We then compute the standardized frequency $\Tilde{f_k}$ for the $k$-th narrative cluster as follows:
\begin{equation}
\label{3}
\tilde{f}_k = \frac{f_k - \mu}{\sigma}
\end{equation}
where $\mu$ is the mean of the frequencies across all clusters and $\sigma$ is the standard deviation of the frequencies. A feature vector $F$ for the document $d$, containing the standardized frequencies of all narrative clusters, can be represented as:
\begin{equation}
\label{4}
F = [\tilde{f}_1, \tilde{f}_2, \ldots, \tilde{f}_k]
\end{equation}

We use this feature vector to train a logistic regression model to predict the article level frames for all $k \in [25, 50 ,75, 100, 125, 150, 175, 200]$. We compare the performance of the narrative chain powered logistic regression model with four baselines. (1) \textbf{Random} - randomly assigns a framing label to each article. (2) \textbf{Latent Dirichlet Allocation (LDA)} - uses LDA \citep{10.5555/944919.944937} with Gibbs Sampling to extract topics from the articles and uses the topic distribution as features to predict the framing labels using a logistic regression model. (3) \textbf{Event Types} - here we evaluate if event types alone can predict framing labels. We use the ETypeClus framework \citep{shen-etal-2021-corpus} to induce event types from the extracted events by clustering the $(verb, object)$ pairs in isolation, without considering any relations. The framework utilizes an expectation-maximization algorithm to simultaneously learn latent event embeddings, as well as learn a latent space with $k$ well-separated clusters. Similar to previous experiments, we use the standardized frequencies of event types in an article as features to predict the framing labels using a logistic regression model. (4) \textbf{Narrative Clusters (w/o LLM expansions)} - instead of using an LLM to expand the narrative chains, we convert them into sentences of the form:

\begin{displayquote}
There is a <causal | temporal> relationship between <$event_1$> and <$event_2$>.
\end{displayquote}
For example:
\begin{displayquote}
There is a causal relationship between (seek, permit) and (pass, legislation).
\end{displayquote}

The rest of the experimental setup remains unchanged.

\begin{table*}[h!]
\centering
\resizebox{\textwidth}{!}{%
\begin{tabular}{cll}
\hline
\multicolumn{1}{l}{\textbf{Frame}} &
  \textbf{Top Ranked Narrative Clusters} &
  \textbf{Narrative Theme} \\ \hline
\multirow{3}{*}{\textbf{\begin{tabular}[c]{@{}l@{}} \\ \\ \\ \\ \\ \\ \\ \\ \\ \\ \\ \\ Crime and Punishment\end{tabular}}} &
  \begin{tabular}[c]{@{}l@{}} \\ \textbf{\emph{((swarm, house), TEMPORAL, (arrest, people))}}: As the officers swarmed the houses, they were \\ able to make the arrests of people suspected of involvement in the smuggling operation.\\ \\ \textbf{\emph{((arrest, Chinese), TEMPORAL, (carry, search))}}: Before the arrest of 22 Chinese individuals, \\ federal agents had been carrying out searches to uncover evidence of their alleged involvement \\ in the scheme.\\ \\ \textbf{\emph{((alarm, investigator), TEMPORAL, (kidnap, criminal))}}: Before investigators were particularly \\ alarmed by groups like the Salvadoran MS-13 gang, they had already been dealing with the reality \\ of kidnappings by criminal suspects.\\ \\ \end{tabular} &
  \begin{tabular}[c]{@{}l@{}}Arresting immigrant smugglers.\end{tabular} \\ \cline{2-3} 
 &
  \begin{tabular}[c]{@{}l@{}} \\ \textbf{\emph{((bring, immigrant), CAUSAL, (find, smuggler))}}: The authorities' ability to find suspected \\ smugglers was directly tied to their efforts to bring undocumented immigrants to shore, where \\ they could be apprehended.\\ \\ \textbf{\emph{((throw, immigrant), CAUSAL, (find, smuggler))}}: The smugglers' decision to throw \\ undocumented immigrants overboard often led authorities to find the smugglers themselves.\\ \\ \textbf{\emph{((fight, drug), TEMPORAL, (deport, worker))}}: As the Mexican gangs continued to fight drug \\ smugglers, the problem of human smuggling from Mexico spilled over into the U.S. Southwest, \\ prompting a growing need to deport workers who were brought into the country illegally. \\ \\ \end{tabular} &
  \begin{tabular}[c]{@{}l@{}}Smuggling of undocumented \\ immigrants.\end{tabular} \\ \cline{2-3} 
 &
  \begin{tabular}[c]{@{}l@{}} \\ \textbf{\emph{((meet, Bush), TEMPORAL, (leave, Mexico))}}: Before meeting with President Bush, \\ President Fox had planned to discuss ways to improve the lives of illegal Mexican immigrants, \\ including finding a documented way for them to leave Mexico.\\ \\ \textbf{\emph{((ask, Bush), TEMPORAL, (grant, Bush))}}: Before asking President Bush to grant amnesty \\ to Mexicans living in the United States, Fox planned to discuss the issue with him.\\ \\ \textbf{\emph{((grant, amnesty), TEMPORAL, (lend, security))}}: Following the announcement that the Bush \\ administration is weighing a plan to grant amnesty to up to 3 million Mexicans, President Fox \\ emphasized the need to lend greater security and orderliness to the migrant flows between Mexico \\ and the United States.\end{tabular} &
  \begin{tabular}[c]{@{}l@{}}Administrations of two countries \\ discussing how to manage the \\ movement of undocumented \\ immigrants across borders. \\ \\ \end{tabular} \\ \\ \hline
\end{tabular}%
}
\caption{Top narrative clusters and their corresponding narrative themes that are strongly predictive of the \emph{Crime and Punishment} frame in the immigration dataset. Narrative chains from each cluster along with their LLM expansions are shown.}
\label{tab:immi-qual}
\end{table*}

Results for these experiments are presented in Fig.~\ref{fig:frame-prediction-results}. We can observe that the narrative cluster model (with LLM expansions) outperforms all four baselines, and that the highest F1 scores are achieved against 150 clusters for the immigration dataset, and 50 clusters for the gun control dataset. The narrative chains obtained through the LLM-guided approach leads to more well defined clusters compared to the non-LLM approach because the former is able to capture richer context from the document as well as more diverse semantic information.

\begin{table}[ht!]
\centering
\resizebox{\columnwidth}{!}{%
\begin{tabular}{cclll}
\hline
\textbf{}                                       & \multicolumn{2}{c}{\textbf{Immigration (k=150)}}                           & \multicolumn{2}{c}{\textbf{Gun Control (k=50)}}                                \\ \hline
                                                & \textbf{Accuracy}                & \multicolumn{1}{c}{\textbf{F1}} & \multicolumn{1}{c}{\textbf{Accuracy}} & \multicolumn{1}{c}{\textbf{F1}} \\ \hline
\textbf{RoBERTa}                                & $0.65_{0.02}$                    & $0.66_{0.02}$                   & $0.65_{0.02}$                         & $0.65_{0.01}$                 \\
\multicolumn{1}{l}{\textbf{+ Narrative Clusters}} & $0.67_{0.03}$ & $0.67_{0.03}$                  & $0.68_{0.01}$                          & $0.66_{0.01}$                 \\ \hline
\end{tabular}%
}
\caption{Accuracy and F1 scores (average and standard deviation) on the frame prediction task for the neural classification model. We trained the model with five different random seeds, and averaged over the results. $k$ is the number of narrative clusters.}
\label{tab:neural-class}
\end{table}

\paragraph{Text + Narrative Cluster Features} Our second experiment combines the cluster features described above with signal from the document text. In this case, we want to show that narrative clusters can introduce significant inductive bias into a simple text classifier, and thus improve performance. 

To do this, we take the best $k$ resulting from the prior experiment for each dataset, and train a neural classifier to predict framing dimensions. Using RoBERTa \citep{liu2019robertarobustlyoptimizedbert}, we obtain a contextualized representation of the entire article and concatenate it with the cluster frequency feature vector. This representation is then passed through a feed-forward net, and the full model is trained end-to-end using the cross entropy loss. Additional implementation details can be found in App.~\ref{sec:neural_model}. 

Results for this experiment are summarized in Tab.~\ref{tab:neural-class}. We observe a minor improvement in performance when we introduce the narrative cluster based feature vectors in the article representations, confirming that this information can indeed introduce inductive bias into the model, and help disambiguate high-level frames. 

The major advantage of our framework is its ability to capture the high level narrative constructs and themes that contribute to framing the different issues in these news articles. Compared to event types and topic clusters which are much more fine-grained in nature, our narrative clusters are able to succinctly capture  high level patterns, thus making these framing dimensions easier to predict. 

\begin{table*}[ht!]
\centering
\resizebox{\textwidth}{!}{%
\begin{tabular}{cll}
\hline
\multicolumn{1}{l}{\textbf{Frame}} &
  \textbf{Top Ranked Narrative Clusters} &
  \textbf{Narrative Theme} \\ \hline
\multirow{3}{*}{\textbf{\begin{tabular}[c]{@{}l@{}} \\ \\ \\ \\ \\ \\ \\ \\ \\ \\ \\  Legality, Constitutionality, \\ Jurisdiction\end{tabular}}} &
  \begin{tabular}[c]{@{}l@{}} \\ \textbf{\emph{((bear, arm), CAUSAL, (keep, arm))}}: The right to bear arms led to the expectation that \\ law-abiding citizens would be allowed to keep their arms.\\ \\ \textbf{\emph{((interpret, Amendment), CAUSAL, (bear, arm))}}: The court's broad interpretation of the \\ Second Amendment led to the conclusion that Americans have a right to bear arms.\\ \\ \textbf{\emph{((bear, arm), TEMPORAL, (protect, right))}}: The Supreme Court's ruling to protect an \\ individual right to keep handguns came after it was established that the Second Amendment \\ allows citizens to bear arms.\end{tabular} &
  Second Amendment right to bear arms. \\ \\ \cline{2-3} 
 &
  \begin{tabular}[c]{@{}l@{}} \\ \textbf{\emph{((reject, ban), TEMPORAL, (protect, right))}}: The court's decision to reject the ban on guns \\ came after it had protected the right to own a firearm.\\ \\ \textbf{\emph{((reconcile, kind), CAUSAL, (ban, possession))}}: The justices' decision to reconcile gun control \\ laws with the Second Amendment was a direct result of their inability to ban the possession of \\ handguns outright.\\ \\ \textbf{\emph{((accept, bar), CAUSAL, (cite, amendment))}}: The state's decision to accept the regulation of \\ handgun ownership led to the district judges citing the amendment in dismissing the cases.\end{tabular} &
  \begin{tabular}[c]{@{}l@{}}Court rulings on constitutionality of \\gun control laws.\end{tabular} \\ \\ \cline{2-3} 
 &
  \begin{tabular}[c]{@{}l@{}} \\ \textbf{\emph{((return, case), CAUSAL, (limit, power))}}: The court's decision to return the case to the lower \\ courts was a direct result of their attempt to limit federal power.\\ \\ \textbf{\emph{((bring, case), TEMPORAL, (hold, unconstitutional))}}: The decision to refuse a rehearing \\ brought the case one step closer to being held unconstitutional.\\ \\ \textbf{\emph{((hear, case), TEMPORAL, (strike, part))}}: After the court refused to revisit the decision to \\ strike down parts of the gun control law, the city's lawyers began evaluating their options \\ to potentially hear the case before the Supreme Court.\end{tabular} &
  \begin{tabular}[c]{@{}l@{}}Courts rejecting appeals in gun control \\cases.\end{tabular} \\ \\ \hline
\end{tabular}%
}
\caption{Top narrative clusters and their corresponding narrative themes that are strongly predictive of the \emph{Legality, Constitutionality, Jurisdiction} frame in the gun control dataset. Narrative chains from each cluster along with their LLM expansions are shown.}
\label{tab:gun-qual}
\end{table*}

\subsubsection{Qualitative Analysis} 

Finally, we perform a qualitative analysis to examine the relation between the resulting narrative clusters and the framing dimensions. To perform this analysis, we compute the mutual information between each narrative cluster and each target frame. This allows us to isolate the narrative clusters that contribute the most towards predicting each frame label. We manually inspect the narrative chains in these clusters to identify high level narrative themes and present partial results in Tab.~\ref{tab:immi-qual} and~\ref{tab:gun-qual}. 

We find that prominent themes supporting the \emph{Crime and Punishment} frame in the immigration dataset talk about ``smuggling of immigrants across the border'' and ``providing amnesty to undocumented immigrants''. Similarly, themes like the ``second amendment right to bear arms'' and ``courts ruling on the constitutionality of gun control laws'' dominate in articles bearing the \emph{Legality, Constitutionality, Jurisdiction} frame from the gun control dataset.

\section{Conclusions and Future Work}

In this paper, we propose a computational framework grounded in event-centric narratives to analyze framing in the news. We used established event extraction methods to construct narrative chains, and adopted an LLM-guided clustering method to capture high level narrative constructs to explain media framing. We performed extensive quantitative and qualitative evaluations of our framework on two different news domains: immigration and gun control. We successfully demonstrated the framework's capability to induce strong thematic narrative clusters that provide significant signal for predicting and explaining the \citet{Boydstun2014} policy frame taxonomy. 

In the future, we would like to: (1) Improve the sub-components of our framework to reduce the noise introduced at different levels, and in turn, improve the quality of the extracted narratives. (2) Explore more effective ways to harness the narrative theme information for predicting and explaining frames. (3) Study the generalizability of our framework for other data sources, domains and framing taxonomies. (4) Employ our framework in a large-scale analysis of framing in the news across time, topics and media outlets. 
\section{Limitations}

The work presented in this paper has three main limitations:

\paragraph{Modeling Complexity and Performance}
This work does not aim to maximize performance headroom with large, complex models, due to the limited computation power we have. Instead, our goal is to highlight a potential research direction for the community, underscoring the importance of identifying key nuances in narrative chains. We aim to stimulate further explorations of this area.

\paragraph{Domain Generalization}
Our method is evaluated and studied for two specific framing datasets: immigration and gun control. The generalization on other topic domains is out of scope of this work and could lead to different conclusions. We save this limitation as an extension in future work.

\paragraph{Narrative Clustering Human Annotation} 
To ensure high-quality evaluation of narrative clustering, two annotators are trained to identify the narrative clustering quality. The annotators possesses full context of this work so are able to engender high quality labels. The average annotation agreement ratio is 79.5\%. However, this annotation quality might not be reproducible through random annotators, or in less popular framing domains.
\section{Ethical Considerations}

To the best of our knowledge, no code of ethics was violated during the development of this project. We used publicly available tools and datasets according to their licensing agreements.

All information needed to replicate our experiments is presented in the paper. We reported all experimental settings, as well as any pre-processing steps, learning configurations, hyper-parameters, and additional technical details. Due to space constraints, some of this information is included in the Appendix.

The analysis reported in Section 4 is done using the outputs of algorithms and machine learning models, and does not represent the authors personal views. The uncertainty of all outputs and predictions was adequately acknowledged in the Limitations section, and the estimated performance was adequately reported. 
\section{Acknowledgement}

This work utilized the Blanca condo computing resource at the University of Colorado Boulder. Blanca is jointly funded by computing users and the University of Colorado Boulder.

\bibliography{./latex/custom}

\appendix
\section{Event Extraction} \label{sec:appendix}
We use the ETypeClus framework \citep{shen-etal-2021-corpus} to extract events. All implementation details can be found in their paper. We replicate all of the hyperparameters from their work. The weight for the clustering-promoting objective is $\lambda = 0.02$, the convergence threshold is $\gamma = 0.05$, and the maximum number of iterations is set to 100. The generative model was learnt using an Adam optimizer with learning rate 0.001 and batch size 64. Only the top 80\% salient verb and objects were considered. Latent space dimensions were $d=100$, and likwise we keep all of the hidden layer dimensions at their default values.

\section{Relation Extraction}
\subsection{Dataset Preprocessing} \label{sec:aser_preprocess}
We find that the majority of the events follow the subject-verb-object (s-v-o) pattern. We use spaCy's dependency parser~\footnote{\url{https://spacy.io/api/dependencyparser}} to extract verb-object pairs for both head and tail event phrases. To handle negative verbs, we identify negation markers such as "no," "not," "n't," "never," and "none" in the context. If a verb is negated, we prepend "not" to it (e.g., "not eat") to accurately reflect its meaning.

In scenarios where the parser extracts multiple verbs or objects for an event phrase, we consider all possible combinations of verb-object pairs. Incomplete pairs, where either the verb or object is missing, are discarded to maintain the integrity of the data.

\subsection{Implementation Details} \label{sec:rel_exp_details}
The relation extraction model is built upon the RoBERTa-based architecture using PyTorch Lightning. The core of the model leverages the pre-trained \emph{roberta-base} model from Hugging Face’s transformers library, which outputs contextualized embeddings for the input tokens.

The model architecture includes a custom classification head that processes the concatenated embeddings of key tokens, such as verbs and objects, from the input sentences. Specifically, it has a hidden layer with a ReLU activation function that maps the combined embeddings into a lower-dimensional space of 100 units. The final layer is a linear classifier that outputs logits for the three target relation classes: Temporal, Causal, and None.

Key hyperparameters used in the model are as follows: learning rate is set to $2*10^-5$, number of epochs is 100, batch size is 8, and maximum token length for the input sequences is set to 256. The model is optimized using a weighted cross entropy loss function.

The model utilizes contextualized embeddings from the RoBERTa model. During the forward pass, the hidden states corresponding to specific tokens, such as verbs and objects, are extracted and averaged to form fixed-size representations. The model accounts for both head and tail entities, including cases where non-verbal (nominalized) verbs are present. These embeddings are concatenated along with the [CLS] token's embedding, creating a feature vector that represents the relation between two entities in the input.

Early stopping is implemented to prevent overfitting, with the training process being monitored by validation loss. The early stopping callback is configured with a patience of 3 epochs, meaning that training will halt if the validation loss does not improve for three consecutive epochs.

Additionally, the model checkpointing mechanism saves the best-performing model based on the lowest validation loss, ensuring that the optimal model is preserved for further evaluation.

During training, the optimizer used is AdamW, which is known for its robustness in handling weight decay. A linear learning rate scheduler with warm-up is employed, where the learning rate linearly increases during the initial warm-up phase and then decays linearly for the remainder of the training.



Results are reported in (Tables~\ref{tab:5fold} and ~\ref{tab:mclass}).

\begin{table}[ht!]
    \centering
    \resizebox{\columnwidth}{!}{%
    \begin{tabular}{cccccc}
    \hline
            \textbf{}   & \multicolumn{1}{l}{\textbf{Accuracy}} & \multicolumn{1}{l}{\textbf{Precision}} & \multicolumn{1}{l}{\textbf{Recall}} & \multicolumn{1}{l}{\textbf{Macro F1}} \\ \hline
\textbf{Fold-1}    & 65.30                                  & 57.97                                  & 65.30                                & 59.94                                 \\
\textbf{Fold-2}    & 65.21                                 & 57.75                                  & 65.21                               & 59.88                                 \\
\textbf{Fold-3}    & 64.77                                 & 57.53                                  & 64.77                               & 59.58                                 \\
\textbf{Fold-4}    & 64.48                                 & 57.05                                  & 64.48                               & 58.93                                 \\
\textbf{Fold-5}    & 64.54                      & 57.24                                             & 64.54                      & 59.14                                          \\ \hline
\textbf{Average}   & \textbf{64.86}                        & \textbf{57.51}                        & \textbf{64.86}                      & \textbf{59.49}                       \\
\textbf{Std. Dev.} & \textbf{0.38}                         & \textbf{0.37}                          & \textbf{0.38}                       & \textbf{0.45}                          \\ \hline
    \end{tabular}}
    \caption{5-fold cross-validation results of the multi-class relation prediction model.}
    \label{tab:5fold}
\end{table}

\begin{table}[ht!]
    \centering
    \resizebox{\columnwidth}{!}{%
    \begin{tabular}{lcccc}
    \hline
            \textbf{Relation}   & \multicolumn{1}{c}{\textbf{Precision}} & \multicolumn{1}{c}{\textbf{Recall}} & \multicolumn{1}{c}{\textbf{F1}} \\ \hline
\textbf{Temporal}  & $0.52_{0.008}$       & $0.67_{0.011}$    & $0.59_{0.005}$      \\
\textbf{Causal}    & $0.33_{0.010}$        & $0.57_{0.010}$     & $0.42_{0.008}$      \\
\textbf{None}      & $0.87_{0.004}$       & $0.71_{0.011}$    & $0.78_{0.004}$      \\ \hline
\textbf{Macro Avg}   & $0.57_{0.007}$       & $0.65_{0.010}$     & $0.60_{0.005}$ \\ \hline
    \end{tabular}}
    \caption{Results for the multi-class relation prediction model (average and std. dev. over all five folds).}
    \label{tab:mclass}
\end{table} 


\section{K-Means Clustering}

We obtain SBERT based sentence embeddings for all narrative chain expansions using the \emph{all-MiniLM-L6-v2} model. Cluster centroids are initialized using the \emph{k-means++} algorithm.

\section{Latent Dirichlet Allocation}

We use a Latent Dirichlet Allocation model with Gibbs sampling. We use the term weighting scheme and set the minimum collection frequency of words to 3, and the minimum document frequency of words is set to 0. We also remove the top 5 most common words. We train the models for a minimum of 1000 iterations.

\section{Neural Frame Prediction Classifier}  \label{sec:neural_model}

We use the RoBERTa model to encode news articles, and use the [CLS] token's embedding as the contextualized embedding for the article. All articles are truncated to 512 tokens. The contextualized article embedding is then combined with the cluster frequency vector and is passed to a classification head. The classification head is a simple two layer feed-forward network with dropout and layer normalization. We use a dropout of 0.3 and the output layer dimensions are 64. We train the model with a batch size of 32, for a maximum of 25 epochs with early stopping with validation on a held out set comprising of 10\% of the training set. We use the Adam optimizer with a learning rate of $2*10^-5$. All parameters are updated during training.

\section{LLM Generation}

We used a 16-bit, instruction fine-tuned Llama 3.1 8B model from the Huggingface Hub. Max tokens was set to 4096. Temperature was set to 0.1.

\section{Running Environment} All experiments were either run on an Intel i9-11900H CPU or on a compute cluster with an A100 GPU with 40GB VRAM. In principle, besides the LLM generated narrative chain expansions, all other experiments should be runnable on CPU.

\section{Random Seed}

We exclusively set all random seeds to 42 for all experiments. The neural classification model is trained on five different seeds in multiples of 7, and averaged results are reported.

\section{Narrative Chain Expansion Prompt} \label{sec:prompts}

We prompt the Llama 3.1 8B model in a zero shot setting, and provide it with the complete news article along with a narrative chain. We first provide a system prompt that explains the task in detail. This is followed by a user prompt, where the actual news article and narrative chain is provided. We show the exact prompts used in the following example for reference.

\paragraph{System Prompt} I want you to generate plausible sentences that expand on an event chain from a news article. Events correspond to what we perceive around us and is denoted as a (VERB, OBJECT) pair. The object is the direct object of the verb in a linguistic sense. An example of an event is (arrest, people). The verb and object will correspond to a word in the article and may or may not be in their lemmatized form. An event chain comprises of two events connected by either a causal or temporal relation. It'll be denoted as a tuple as follows: (EVENT\_1, RELATION\_TYPE, EVENT\_2). RELATION\_TYPE can be either CAUSAL or TEMPORAL. CAUSAL indicates that EVENT\_2 occurred as a result of EVENT\_1 or EVENT\_2 is the reason why EVENT\_1 occurred. TEMPORAL indicates EVENT\_2 occurred before, after or synchronously with EVENT\_1. An example of an event chain is ((arrest, people), CAUSAL, (protest, legislation)). I will provide you with an event chain and the corresponding news article to which it belongs. I want you to expand the event chain into a plausible sentence.

\paragraph{User Prompt} News Article: <Full Text of the News Article>. Event Chain: ((reject, ban), TEMPORAL, (protect, right)). Generate a very short sentence that expands the events in the event chain and the relationship between them in the context of the news article. Do not generate anything else. \\

For the sake of brevity, we used a placeholder for the news article in the example user prompt.

\end{document}